# RBIR Based on Signature Graph

Thanh The Van
Center for Information Technology
HCMC University of Food Industry
HoChiMinh, Vietnam
thanhvt@cntp.edu.vn

Thanh Manh Le
Hue University
Hue, Vietnam
lmthanh@hueuni.edu.vn

*Abstract*—**This paper approaches the image retrieval system on the base of visual features local region RBIR (region-based image retrieval). First of all, the paper presents a method for extracting the interest points based on Harris-Laplace to create the feature region of the image. Next, in order to reduce the storage space and speed up query image, the paper builds the binary signature structure to describe the visual content of image. Based on the image's binary signature, the paper builds the SG (signature graph) to classify and store image's binary signatures. Since then, the paper builds the image retrieval algorithm on SG through the similar measure EMD (earth mover's distance) between the image's binary signatures. Last but not least, the paper gives an image retrieval model RBIR, experiments and assesses the image retrieval method on Corel image database over 10,000 images.**

***Keywords-Image Retrieval; Binary Signature; Signature Graph***

## I. INTRODUCTION

There are three common approaches for query image [1], including: image retrieval based on TBIR key word (text-based image retrieval), image retrieval based on CBIR content (content-based image retrieval) and image retrieval based on semantic SBIR (semantic-based image retrieval). The image retrieval through TBIR key word will be difficult and time-consuming in the description of image's content. Thus, it is necessary to build CBIR retrieval system through the content of image to find out the similar image. Furthermore, when querying image through a key word or an index, we will not describe visual characteristics of image. So, we need to create a method of extracting image's characteristics, thence reflect the content of image to find out images similar content. Extracting visual features of image is an important task of the image retrieval process on the base of content. However, if we query and compare directly the content of image, the problem will be complicated, time-consuming and costly storage space. For this reason, when comparing the image's content, still ensuring the query speed and storage space.

A number of works related to the query image's content has been published recently as: Extracting image objects based on the change of histogram value [1], Similarity image retrieval based on the comparison of characteristic regions and the similar relationship of feature regions on image [2], Color image retrieval based on the detection of local feature regions by Harris-Laplace [3], Color image retrieval based on bit plane and Lab color space [4], Converting color space and building hash table in order to query the content of color images [5],...

This paper approaches the semantic description of image's content through a binary signature and builds data structure to store this one. This data structure describes the relationship between the binary signatures, which describes the relationship between image's contents. Based on the description of the semantic relationship of this data structure's image content, the paper will find out the similar image in content on Corel image database [6].

## II. THE SIMILAR MEASURE

### A. *Image's Binary Signature*

In accordance [7], the binary signature is formed by hashing the data objects, and will have $k$ bit 1 and $(m-k)$ bit 0 in the bit chain $[1..m]$, with $m$ is the length of the binary signature [7]. The data objects and the object of the query are encoded on the same algorithm. When the bits in the signature data object are completely covered with the bits in the query signature, then this data object is a candidate fulfilling the query. According to [7] there are three cases: (1) the data object matches the query: every bit in the $s_q$ is covered with the bits in the signature $s_i$ of the data object (i.e., $s_q \wedge s_i = s_q$); (2) the object does not match the query (i.e., $s_q \wedge s_i \neq s_q$); (3) the signatures are compared and then give the *false drop* result.

***Definition 1:*** Setting $F = (F_1, F_2, ..., F_{n_F})$ is a vector to describe the feature regions of image. Setting vector $F(R_i^I) = (F_1(R_i^I), F_2(R_i^I), ..., F_{n_F}(R_i^I))$ is a vector value of feature regions which was standardized on [0,1] (i.e. $F_j(R_i^I) \in [0,1]$ , $j = 1,...,n_F$ ). Setting $B_I^j = b_1^j b_2^j ... b_m^j$ with $b_k^j = 1$ if $k = \left[ (F_j(R_i^I) + 0.05) \times m \right]$ , otherwise $b_k^j = 0$ ( $k = 1,...,m$ ). At that time, the binary signature of feature region $R_i^I$ be defined as: $Sig(R_i^I) = B_I^1 B_I^2 ... B_I^{nF}$ . The binary signature of image $I$ will be $Sig(I) = Sig(R^I) = \bigcup_i Sig(R_i^I)$ .

### B. *EMD Distance*

Setting $I$ is a set of suppliers, $J$ is a set of consumers, $c_{ij}$ is the transportation cost from the supplier $i \in I$ to the consumer $j \in J$ , we need to find out flows $f_{ij}$ to minimize the total cost $\sum_{i\in I}\sum_{j\in J} c_{ij} f_{ij}$ with the constraints [8]: $f_{ij} \ge 0, \sum_{i\in I} f_{ij} \le y_j, \sum_{j\in J} f_{ij} \le x_i, i\in I, j\in J$ . With $x_i$ is the provider's general ability $i\in I$ , $y_j$ is the total need of the consumer $j\in J$ . The feasible condition is $\sum_{j\in J} y_j \le \sum_{i\in I} x_i$ . The EMD distance [8] as follow:

$$EMD(x,y)=(\sum_{i\in I}\sum_{j\in J} c_{ij} f_{ij})/(\sum_{i\in I}\sum_{j\in J} f_{ij})=(\sum_{i\in I}\sum_{j\in J} c_{ij} f_{ij})/(\sum_{j\in J} y_j)$$

The image $I$ with the binary signature is $SIG_I = B_I^1 B_I^2 ... B_I^n$ , with $B_I^j = b_1^j b_2^j ... b_m^j$, $b_i^j \in \{0,1\}$. The weight of $B_I^j$ component as: $w_I^j = w(B_I^j) = \sum_{i=1}^{m}(b_i^j \times \frac{i}{m} \times 100)$ . Therefore, we have the weight vector of image $I$ will be $W_I = \{w_I^1, w_I^2, ..., w_I^n\}$ . Setting $J$ is the image needs to calculate the similarity corresponding to the image $I$ , so we need to minimize the cost of converting color distribution is as: $\sum_{i=1}^{n}\sum_{j=1}^{n} d_{ij} f_{ij}$ , with $F = (f_{ij})$ as a matrix of color distribution flows between $c_I^i$ and $c_J^j$, and $D = (d_{ij})$ as a matrix of Euclidean distance in RGB color space between $c_I^i$ and $c_J^j$ . Setting $W_m = \min(\sum_{i=1}^{n} w_I^i, \sum_{j=1}^{n} w_J^j)$ and $W_M = \max(\sum_{i=1}^{n} w_I^i, \sum_{j=1}^{n} w_J^j)$, then the distance $EMD(I,J)$ is a color distribution flows from the image with the color weight $W_M$ to the image with color weight $W_m$ . For this reason, the similar measure between two images $I$ and $J$ on the base of EMD distance will be the minimum cost:

$$EMD(I,J) = \min_{F=(f_{ij})} \frac{(\sum_{i=1}^{n}\sum_{j=1}^{n} d_{ij} f_{ij})}{\sum_{i=1}^{n}\sum_{j=1}^{n} f_{ij}}, \text{ with } \sum_{i=1}^{n}\sum_{j=1}^{n} f_{ij} = W_m = \min(\sum_{i=1}^{n} w_I^i, \sum_{j=1}^{n} w_J^j)$$

## III. DATA STRUCTURE AND IMAGE RETRIEVAL ALGORITHM

### A. *The Signature Graph*

***Definition 2:*** The Signature Graph $SG = (V, E)$ is the graph which describes the relationship between the image, has a set of vertices $V = \{\langle oid_I, Sig(R^I)\rangle \mid I \in \Im\}$ and a set of edges $E = \{\langle I, J\rangle \mid \phi(I,J) = \phi(R^I, R^J) \le \theta(R^I, R^J),\ \forall I, J \in \Im\}$. With the weight of each edge $\langle I, J\rangle$ is a measurement function of the similarity $\phi(I,J) = \phi(R^I, R^J)$ , $\theta(R^I, R^J)$ is a threshold value and $\Im$ is an image database. In this paper, the similar measure $\phi(I,J) = EMD(I,J) = EMD(Sig(I), Sig(J))$ .

Each vertex $v \in V$ in $SG$ will determine $k$ element which has the nearest similar measurement. However, if the number of image in a large database, we will be difficult to determine the set of similar image corresponding to the query image. Therefore, we build the notion of S-*k*Graph so that every vertex include $k$ nearest image and call as *k-neighboring* image.

***Definition 3:*** A cluster $V_i$ has center $I_i$ , with $k_i\theta$ is a radius, be defined as follow:

$$V_i = V_i(I_i) = \{J \mid \phi(I_i, J) \le k_i\theta, J \in \Im,\ i = 1,...,n\},\ k_i \in N^*.$$

***Definition 4:*** *(S-kGraph)* Giving a set $\Omega = \{V_i \mid i = 1,...,n\}$ is a set of cluster, with $V_i \cap V_j = \emptyset, i \ne j$ . The S-*k*Graph = $(V_{SG}, E_{SG})$ is the graph which have the weight, including vertex set $V_{SG}$ and edge set $E_{SG}$ be defined as follow:

$$V_{SG} = \Omega = \{V_i \mid \exists! I_{i_0} \in V_i, \forall I \in V_i, \phi(I_{i_0}, I) \le k_{i_0}\theta, i = 1,...,n\}$$

$$E_{SG} = \{\langle V_i, V_j\rangle \mid i \ne j, V_i \in V_{SG}, V_j \in V_{SG}, d(V_i, V_j) = \phi(I_{i_0}, J_{j0})\},$$

with $d(V_i, V_j)$ is the weight between two clusters and $\forall I \in V_i, \phi(I_{i_0}, I) \le k_{i_0}\theta$ .

***Definition 5:*** *(The Rules of Distribution of Image)* Giving set $\Omega = \{V_i \mid i = 1,...,n\}$ is a set of clusters, with $V_i \cap V_j = \emptyset, i \ne j$, setting $I_0$ is an image needs to distribute in a set of clusters $\Omega$ , setting $I_m$ is a center of cluster $V_m$ so that $(\phi(I_0, I_m) - k_m\theta) = \min\{(\phi(I_0, I_i) - k_i\theta), i = 1,...,n\}$ , with $I_i$ is a center of cluster $V_i$ . There are three cases as follow:

(1) If $\phi(I_0, I_m) \le k_m\theta$ then the image $I_0$ will be distributed in cluster $V_m$ .

(2) If $\phi(I_0, I_m) > k_m\theta$ then setting $k_0 = \lceil(\phi(I_0, I_m) - k_m\theta)/\theta\rceil$, at that time:

(2.1) If $k_0 > 0$ then creating cluster $V_0$ with center $I_0$ and radius is $k_0\theta$ , at that time $\Omega = \Omega \cup \{V_0\}$ .

(2.2) Otherwise (i.e. $k_0 = 0$ ), the image $I_0$ will be distributed in cluster $V_m$ and $\phi(I_0, I_m) = k_m\theta$ .

### B. *Extracting the Feature Region*

For extracting the visual feature of image, the first step is standardized the image size (i.e. convert input image which has different size into the image have size $k \times k$ ) , from that extracting the color feature of image. Because the image according to JPEG standard is described on color space YCbCr, so we need use YCbCr to extract specific information of the image. Setting Y, Cb, Cr as a intensity, Blue color, Red color respectively. In accordance [5], the convertibility from RGB to YCbCr as follow (with $R, G, B \in [0,1]$ ):

$$\begin{bmatrix} Y \\ Cb \\ Cr \end{bmatrix} = \begin{bmatrix} 65.481 & 128.553 & 24.996 \\ -37.797 & -74.203 & 112 \\ 112 & -93.786 & -18.214 \end{bmatrix} \begin{bmatrix} R \\ G \\ B \end{bmatrix} + \begin{bmatrix} 16 \\ 128 \\ 128 \end{bmatrix}$$

According to [3], [4], the Gaussian transformation by human's visual system is as follow:

$$L(x,y,\delta_D) = \frac{1}{10}[6.G(x,y,\delta_D)*Y + 2.G(x,y,\delta_D)*Cb + 2.G(x,y,\delta_D)*Cr]$$

with $G(x,y,\delta_D)=\frac{1}{\sqrt{2\pi}.\delta_D}.\exp(\frac{x^2+y^2}{2.\delta_D^2})$

The intensity $I_0(x,y)$ for color image is calculated according to equation:

$I_0(x,y,\delta_I,\delta_D)=Det(M(x,y,\delta_I,\delta_D))-\alpha.Tr^2(M(x,y,\delta_I,\delta_D))$

In there, $Det(\bullet),Tr(\bullet)$ as *Determinant* and *Trace* of matrix, $M(x,y,\delta_I,\delta_D)$ is a second moment matrix, be defined as follow:

$$M(x,y,\delta_I,\delta_D)=\delta_D^2.G(\delta_I)*\begin{bmatrix} L_x{}^2 & L_xL_y \\ L_xL_y & L_y{}^2 \end{bmatrix}$$

In there, $\delta_I,\delta_D$ are the integration scale and differentiation scale, and $L_\alpha$ is the derivative computed the $\alpha$ direction. The feature points of color image are extracted according to formula: $I_0(x,y,\delta_I,\delta_D)>I_0(x',y',\delta_I,\delta_D)$, with $x',y'\in A$

$$I_0(x,y,\delta_I,\delta_D)\geq\theta$$

with $A$ is the neighborhood of point $(x,y)$ and $\theta$ is a threshold value. A set of feature circles $O_I=\{o_I^1,o_I^2,...,o_I^n\}$ has center is a feature points and a set of feature radius $R_I=\{r_I^1,r_I^2,...,r_I^n\}$. Values of feature radius are extracted according to LoG method (Laplace-of-Gaussian) and have value in $[0,\min(M,N)/2]$, with $M,N$ are the height and the width of image.

### C. Creating a Binary Signature of the Image

Each feature region $o_I^i\in O_I$ of the image $I$ will be calculated histogram on the base of the standard color range $C$, effectuating the clustering method relies on Euclidean measure in RGB color space to classify colors of every pixel on the image. Setting $p$ is a pixel of image $I$ and has a color vector in RGB as $V_p=\left(R_p,\ G_p,\ B_p\right)$. Setting $V_m=\left(R_m,\ G_m,\ B_m\right)$ is a color vector of a set of standard color range $C$, so as to: $V_m=\min\{\|V_p-V_i\|,\ V_i\in C\}$. At that time, at pixel $p$ will be standardized in accordance with color vector $V_m$. According to experiment, the paper will use the standard color range on MPEG7 to calculate histogram for color images on Corel database.

Setting $o_I^i\in O_I$ ($i=1,...,N$) is a feature circle of the image $I$, histogram vector of the circle $o_I^i$ according to MPEG7 standard will be as: $H(o_I^i)=\{H_1(o_I^i),H_2(o_I^i),...,H_n(o_I^i)\}$, setting $h_k(o_I^i)=H_k(o_I^i)\Big/\sum_j H_j(o_I^i)$, standardize histogram vector is as $h(o_I^i)=\{h_1(o_I^i),h_2(o_I^i),...,h_n(o_I^i)\}$. When, the binary signature to describe $h_k(o_I^i)$ will be as $B_I^k=b_I^1b_I^2...b_I^m$, with $b_I^j=1$ if $j=\left[(h_j(o_I^i)+0.05)\times m\right]$, otherwise $b_I^j=0$. So, the signature to describe the feature region $o_I^i\in O_I$ as $Sig(o_i^I)=B_I^1B_I^2...B_I^n$. For this reason, the binary signature of the image $I$ will be as: $S_I=\bigcup_{i=1}^N Sig(o_I^i)$

### D. Creating S-kGraph

**Input:** Image database $\Im$ and threshold $k\theta$
**Output:** S-kGraph = $(V_{SG},E_{SG})$
***Algorithm1.*** Create_S-kGraph($\Im$, $k\theta$)
**Begin**
$V_{SG}=\varnothing;\ \ E_{SG}=\varnothing;\ k_I=1;\ n=1;$
**For** ($\forall I\in\Im$) **do**
**Begin**
**If** ($V_{SG}=\varnothing$) **then**
  **Begin**
  $I_0^n=I;\qquad r=k_I\theta;$
  Initialize cluster $V_n=\langle I_0^n,r,\phi=0\rangle;$
  $V_{SG}=V_{SG}\cup V_n;$
  **End**
**Else**
  **Begin**
    $(\phi(I,I_0^m)-k_m\theta)=\min\{(\phi(I,I_0^i)-k_i\theta),i=1,...,n\}$;
    **If** ($\phi(I,I_0^m)\leq k_m\theta$) **then**
      $V_m=V_m\cup\langle I,k_m\theta,\phi(I,I_0^m)\rangle;$
    **Else**
    **Begin**
    $k_I=\left[(\phi(I,I_0^m)-k_m\theta)/\theta\right];$
    **If** ($k_I>0$) **then**
    **Begin**
    $I_0^{n+1}=I;\ r=k_I\theta;$
    Initialize cluster $V_{n+1}=\langle I_0^{n+1},r,\phi=0\rangle;$
    $V_{SG}=V_{SG}\cup V_{n+1};$
    $E_{SG}=E_{SG}\cup\{\langle V_{n+1},V_i\rangle\mid\phi(I_0^{n+1},I_0^i)\leq k\theta,i=1,...,n\};$
    $n=n+1;$
    **End**
    **Else**
      **Begin**
        $\phi(I,I_0^m)=k_m\theta;$
        $V_m=V_m\cup\langle I,k_m\theta,\phi(I,I_0^m)\rangle;$
      **End**
    **End**
**End**
**End.**

### E. Image Retrieval Algorithm

**Input:** query image $I_Q$, S-$k$Graph = $(V_{SG},E_{SG})$, threshold $k\theta$.
**Output:** set of similar image IMG.
***Algorithm2.*** Search-Image($I_Q$, S-$k$Graph, $k\theta$)
**Begin**

IMG = $\varnothing$;

$V = \varnothing$; $\phi_{\min} = \phi(I_Q, I_0^m) = \min\{\phi(I_Q, I_0^i), i = 1,...,n\}$;

**For** ( $V_i \in V_{SG}$ ) **do**

**If** ( $\phi(I_0^m, I_0^i) \le k\theta$ ) **then** $V = V \cup V_i$ ;

**For** ( $V_j \in V$ ) **do**

IMG = IMG $\cup$ $\{I_k^j, I_k^j \in V_j, k = 1,...,|V_j|\}$ ;

**return** IMG;

**End.**

## IV. EXPERIMENTS

### A. *Model of Region-Based Image Retrieval*

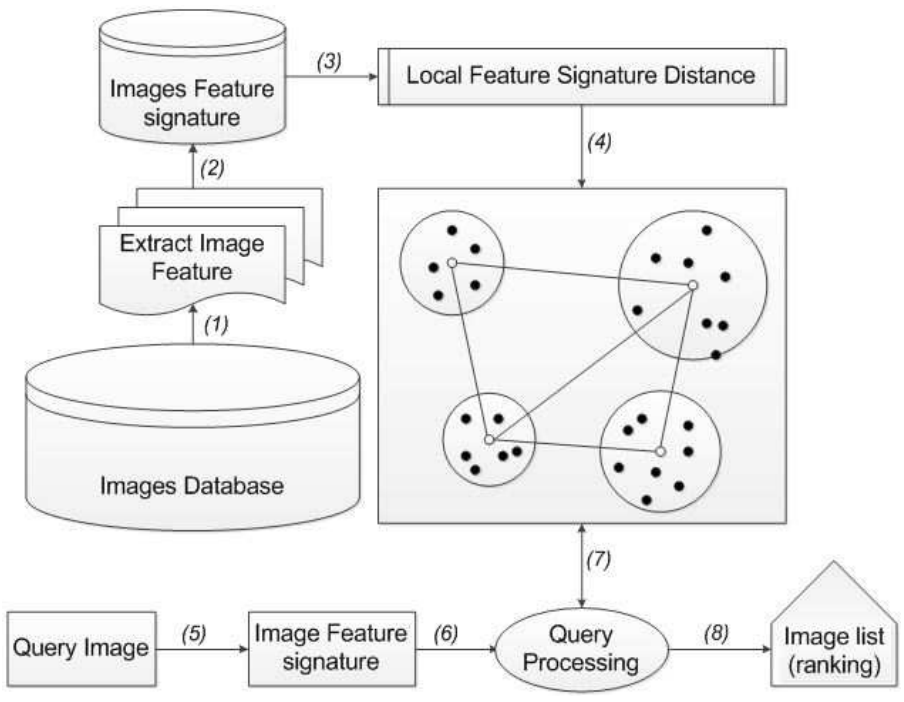

Figure 1. The model of RBIR based on signature graph

**Phase 1:** *Perform pre-processing*

*Step 1.* Extract feature regions of the image in database into feature vector.

*Step 2.* Convert the feature vector of the image in the form of binary signature.

*Step 3.* Calculate the EMD distance of the feature signature of the image and insert into S-*k*Graph.

**Phase 2:** *Implement Query*

*Step 1.* For each query image, will extract the feature vector and convert into binary signature.

*Step 2.* Perform the process of binary signature retrieval on S-*k*Graph to find out the similar image.

*Step 3.* After we have the similar images, carry out an arrangement according to similarity from high to low and give the list of the image based on the similarity of binary signature.

### B. *The Experimental Results*

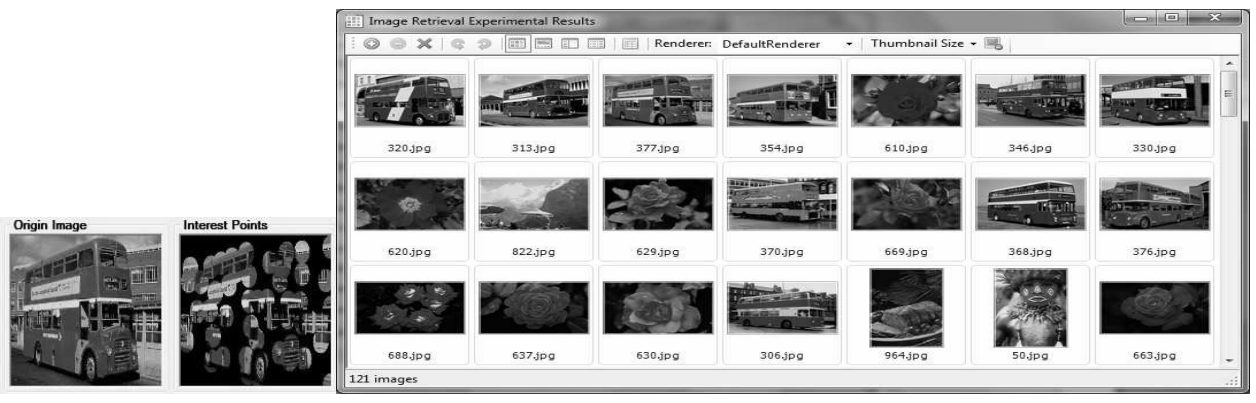

Figure 2. A sample result about image retrieval based on signature graph.

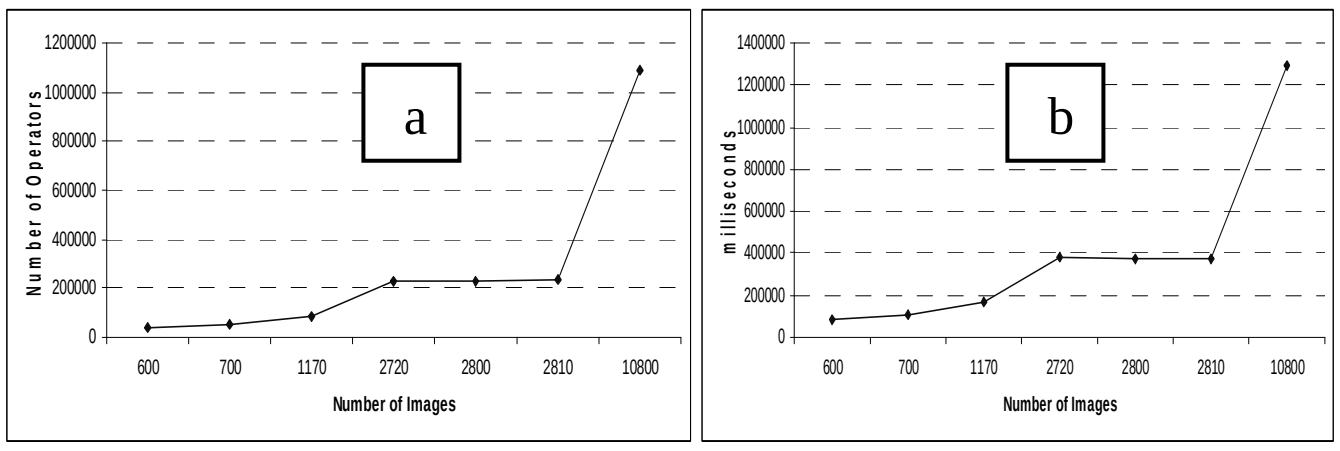

Figure 3. a- Number of comparisons to create SG. b- The time to create SG

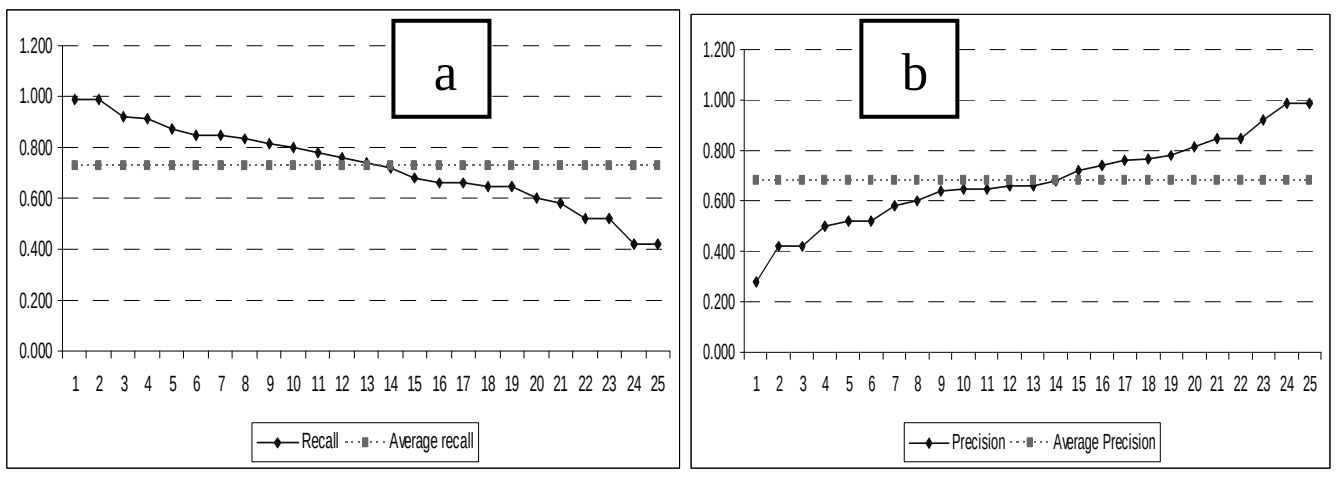

Figure 4. a- Recall. b- Precision

## V. CONCLUSION

The paper gives the similarity evaluation method between two images on the base of binary signature and creates the S-*k*Graph to describe the relationship in of images. Since then, the paper creates the image retrieval system model on the base of feature regions, at that time to simulate the experiment on Corel's image data classification. According to experimental result, the method of evaluation based on S-*k*Graph has speed up query similar images more than query in SSF (sequential signature file). However, the use of the feature of color will give an inaccurate result in the sense of image content. Therefore, the next development of the paper will extract the object on the image, since the paper will build the binary signature to describe objects as well as to describe the content of image.